\documentclass[letterpaper, 10 pt, conference]{ieeeconf}  

\IEEEoverridecommandlockouts                              

\usepackage[dvipdfmx]{graphicx}
\usepackage{color}

\title{\LARGE \bf
Learning Bimanual Manipulation via Action Chunking and Inter-Arm Coordination with Transformers
}

\author{Tomohiro Motoda$^{1, \dagger}$, Ryo Hanai$^{1}$, Ryoichi Nakajo$^{1}$, Masaki Murooka$^{1}$, Floris Erich$^{1}$ and Yukiyasu Domae$^{1}$
\thanks{$^\dagger$Corresponding author, reachable: {tomohiro.motoda@aist.go.jp}
}
\thanks{$^{1}$National Institute of Advanced Industrial Science and Technology (AIST), Tokyo, Japan.
}%
}

\begin{document}

\maketitle
\thispagestyle{empty}
\pagestyle{empty}

\begin{abstract}
Robots that can operate autonomously in a human living environment are necessary to have the ability to handle various tasks flexibly. One crucial element is coordinated bimanual movements that enable functions that are difficult to perform with one hand alone. In recent years, learning-based models that focus on the possibilities of bimanual movements have been proposed. However, the high degree of freedom of the robot makes it challenging to reason about control, and the left and right robot arms need to adjust their actions depending on the situation, making it difficult to realize more dexterous tasks. To address the issue, we focus on coordination and efficiency between both arms, particularly for synchronized actions. Therefore, we propose a novel imitation learning architecture that predicts cooperative actions. We differentiate the architecture for both arms and add an intermediate encoder layer, Inter-Arm Coordinated transformer Encoder (IACE), that facilitates synchronization and temporal alignment to ensure smooth and coordinated actions. To verify the effectiveness of our architectures, we perform distinctive bimanual tasks. The experimental results showed that our model demonstrated a high success rate for comparison and suggested a suitable architecture for the policy learning of bimanual manipulation.
\end{abstract}

\section{INTRODUCTION}

Humans demonstrate exceptional dexterity, particularly in tasks requiring high precision, such as complex assembly operations. Bimanual manipulation enables efficient simultaneous execution of multiple picking tasks. For instance, one arm can stabilize an object while the other inserts screws in assembly lines, or both arms can lift and manipulate larger items together.

Integrating multiple robotic systems facilitates tasks that may prove challenging for single-arm robots by implementing coordinated bimanual operations. Advances in foundation models leveraging extensive datasets and imitation learning techniques have driven substantial progress in this field. Noteworthy systems like ALOHA~\cite{zhao2023learning} have successfully utilized data from dual-arm robots and improved bimanual control by employing well-designed architectures. 
Nevertheless, utilizing a solitary, trained model for controlling robots gives rise to many challenges, which can be attributed to variations in hardware configurations. These challenges stem from architectural limitations and differences in the physical and action spaces across the robots, which contribute to data heterogeneity. For instance, deploying a bimanual robot with 12 or 14 degrees of freedom (DoF) using two standard single-arm robots, each with 6 DoF—such as the ViperX-300, UR5, or 7 DoF as the Franka, still necessitates fine-tuning or training the model from scratch. This paper proposes an approach that further advances bimanual manipulation by refining the coordination between both arms through enhanced action synchronization and skill acquisition processes.

Our key insight is to examine the optimal architectural design that supports learning bimanual actions in a twin-armed task, ensuring that each arm of the robot acquires basic skills. For instance, in the common bimanual operation of transferring objects, the two-step task involves using the right hand to grasp an object while the left hand receives it, reflecting the distinct roles of the left and right arms. We hypothesize that deep single-arm learning is important for learning dual-arm manipulation. Our findings provide a clear perspective on introducing an architecture with a path focused on a single arm versus relying on end-to-end learning without considering the structure of both arms and demonstrate the benefits of evaluating the relationship between hardware and software.

We propose a novel architecture,  the Inter-Arm Coordinated transformer Encoder (IACE), that can adjust the synchronization and timing of potential bimanual movements against the encoders corresponding to each arm. Our overall model contains a local Transformer encoder for each robot arm trajectory, the IACE  to facilitate learning bimanual actions, and a Transformer decoder to infer the action chunk. We compare two types of Transformer decoders: split decoders and single decoders. 

Overall, we emphasize that our architecture significantly advances bimanual robot manipulation. Building on these insights, we make the following contributions: 
\begin{itemize}
    \item The present study investigates the most appropriate robot model architectures based on an essential combination of encoders and decoders for deploying bimanual robots. The proposed architecture incorporates an internal encoder for the left and right arms to ensure synchronized operation and enhance the control of the bimanual robots. 
    \item The proposed architecture includes the Inter-Arm Coordinated transformer Encoder (IACE) layer, which features a mechanism within the encoder to synchronize and coordinate the temporal dynamics of the movements of both arms. The actions performed by each hand depend on the movements of both arms. To this end, the architecture enables coordinated control of the left and right arm movements. To verify our proposed model, we compare it with multiple baselines and conduct preliminary validation with cases that do not include IACE to confirm its effectiveness.
    \item We conduct distinctive bimanual tasks to show that our model has an 8-9\% higher success rate than ACT~\cite{zhao2023learning}. 

\end{itemize}

\begin{figure*}[t]
        \centering
        \includegraphics[width=\linewidth]{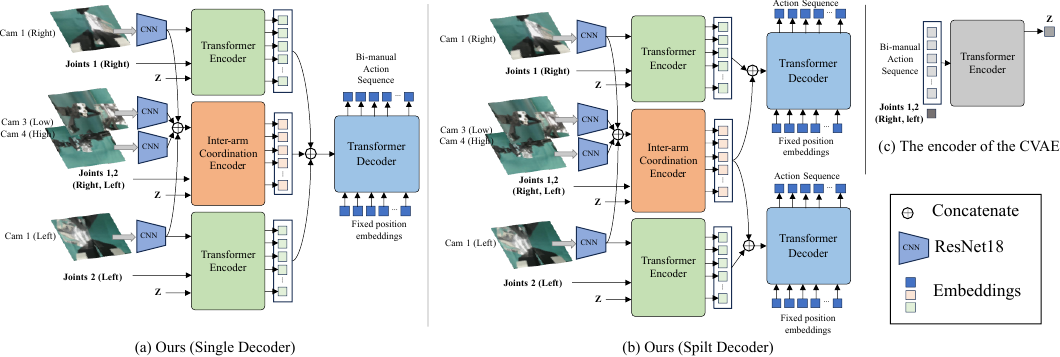}
        \vspace{-15pt}
        \caption{\textbf{Details of Our proposed architectures.} (a) Single Decoder to process information from the multiple encoders, simplifying the decoding process. (b) Split Decoders get the input from each encoder, process, and output each action. It allows each decoder to handle different parts of the data, potentially improving performance on complex bimanual tasks (c) CVAE Encoder is crucial for reconstructing inputs or generating new samples to eliminate the variation of human teleoperation data. }
        \label{fig:architectures}
\end{figure*}

\section{Related works}
    \subsection{Imitation learning for robotic manipulation.}
    The development of these robotic learning models relies on two key components: (1) the collection of high-quality trajectory, image, and knowledge data from both real-world and simulated environments~\cite{openxembodiment}, and (2) the development of flexible architectures that can effectively utilize episodic data for ongoing improvement. Controlling robots with diverse morphologies remains challenging due to variations in hardware configurations, such as differences in shape and layout. Researchers are exploring innovative approaches, including adaptable architectures like foundation models to address this. This ongoing research holds great promise for the future of robotics.
    
    Behavioral Cloning (BC) is one of the imitation learning algorithms that replicates actions from collected data~\cite{Pomerleau-1989-15721}. Recent Behavioral Cloning advancements have leveraged extensive robot learning and significant progress. For instance, BC has been utilized to achieve actual robot actions from one-shot imitation~\cite{duan2017one, yu2018oneshothuman}, and to develop multitasking capabilities through language~\cite{Bharadhwa2024jRoboAgent}. Additionally, BC has been applied to Task and Motion Planning (TAMP) agents~\cite{dalal2023optimus}, and bilateral control~\cite{Buamanee2024biact}. As well, Transformer~\cite{vaswani2017transformer} has leveraged many recent studies, e.g., decision-making using ~\cite{chen2021decision} and fine-grained action prediction through Action Chunking~\cite{zhao2023learning}. Diffusion processes have been employed in robot action generation; Diffusion Policy~\cite{chi2023diffusionpolicy} creates diverse and precise trajectories from the small dataset. Notable examples include large-scale behavior inference models~\cite{brohan2022rt1}, which have constructed extensive databases with practical applications in diverse scenarios. Other research has introduced frameworks inspired by human brain mechanisms, known as deep predictive learning~\cite{suzuki2023deep, Friston2005}. These frameworks utilize recurrent neural networks (RNNs) to minimize prediction errors between sensory-motor information at the current and subsequent steps, leading to robust motion generation. This has been demonstrated in practical scenarios, particularly for flexible objects that are challenging to model, such as folding cloth~\cite{Yang2017RepeatableFolding} and opening doors~\cite{ito2022Door}, adapting to object states. Studies focusing on trajectory prediction have examined the information of environments and agents, including robots~\cite{Wang2021MultiPerson3M, valls2024echo}. These studies involve learning from human actions to predict appropriate robot movements within various scenes~\cite{kedia2024interact}.
    Our research aims to improve learning architectures for dual-arm robots by developing optimal behavioral cloning algorithms that adapt to the differences and objectives of dual-arm movements.

    \subsection{Frameworks for Learning Bimanual Manipulation}

    Bimanual manipulation is highly demanded across various operations and is a crucial core in dual-arm manipulator and humanoid research~\cite{SMITH20121340, Abbas2023}. Recent developments in dual-arm robot systems, such as ALOHA~\cite{zhao2023learning}, MobileALOHA~\cite{fu2024mobile}, and BiDex~\cite{shaw2024bimanual}, have demonstrated the capability to perform tasks in a human-like manner. Bimanual manipulation is especially significant for performing practical tasks in everyday life. For example, Kim et al.~\cite{kim2024dual} have focused on policy learning using large-scale data for precise dual-arm tasks. Large foundation models like $\pi0$~\cite{black2024pi0visionlanguageactionflowmodel} have also incorporated extensive bimanual robot data for various tasks. Hardware differences in robots significantly influence the learning policy trained from various datasets. Thus, data related to bimanual robots require advancements in architecture and learning algorithms, especially when mixed with single-arm models. By incorporating a certain degree of flexibility in the architectures, it has been shown that a single architecture can accommodate both dual-arm and single-arm models by combining multiple action heads~\cite{octo2024, doshi24-crossformer} and generalizing the action space~\cite{liu2024rdt1b}. Lu et al.~\cite{lu2024anybimanual} have extended dual-arm tasks by managing the skills of each arm. In our work, we assume that dual-arm robots possess two arms and consider effective architectures for single-arm learning to achieve appropriate trajectories. 
    
    Hierarchical mechanisms are known to be effective in robot learning~\cite{Yamashita2008Hierarchy}. Recent models incorporating attention mechanisms benefit from handling multimodal inputs, and hierarchical attention transformers~\cite{lee2024interactinterdependencyawareaction} have been utilized to synchronize the trajectories of both arms, enhancing dual-arm task learning. Shikada et al.~\cite{shikada2024bimanual} proposed a hierarchical mechanism. The Deep Predictive Learning Model focused on bimanual tasks, demonstrating the synchronization of robot trajectories for the right and left arms through hierarchical layers. We introduce an architecture with encoder layers for each arm to sufficiently learn the movements and a global layer to grasp the overall motion and context, thereby learning the timing of bilateral movements.
    
\section{Action Chunking and Inter-Arm Coordination with Transformers}

    It is crucial to develop architectures that improve data efficiency and task success to enhance the effectiveness of policy learning for various robot models using limited data. However, as in this study, doubling the input-output dimensions is insufficient when addressing bimanual manipulation. This is because specific tasks can only be accomplished through the coordinated efforts of both arms, which wouldn't be achievable with just one arm. Therefore, our primary goal is to create architectures that enhance imitation learning for bimanual manipulation and to investigate the characteristics of such learning through thorough validation of these configurations.

    \subsection{Network Architecture}

    We build our proposed models on the ACT model to design different encoder and decoder structures. In particular, we propose a new design called the inter-arm coordinated transformer Encoder (IACE), which helps synchronize and time the movements of both arms. 

    We propose basic architectures that consist of encoders and decoders assigned to each arm, designed to leverage the bimanual manipulation. Our approach features the IACE, allowing the individual robot arms to learn their trajectories while simultaneously considering the state of the other arm during bimanual tasks. This enables stable inference of the bimanual robot trajectories. The network architectures are illustrated in Fig.~\ref{fig:architectures}. 

    We combined the joint sequences from both arms with the visual feature sequence to create a comprehensive input sequence. To help draw attention to and utilize information from each segment, we included classification (CLS) tokens at the beginning of each segment~\cite{devlin2019bertpretrainingdeepbidirectional}. Additionally, we applied positional embeddings to the sequence to ensure that the model understands the order of the tokens within each segment. 
    
    The model should focus on the corresponding wrist camera and joint values to determine the appropriate trajectory for each robot arm. Each arm is supported by its local encoder. Global information is also integrated to account for the interdependence of left and right movements in dual-arm tasks. This allows the model to model synchronization and timing during movement effectively. We call it the Inter-Arm Coordinated transformer Encoder.   In the decoder, we considered that it is necessary to verify to what extent batch output or split output is adequate for the twin-arm task. Therefore, in this study, two models, (a) single decoder and (b) split decoders, are applied in the decoder setup, as shown in Fig. 2. Details of the architecture of each component are queried below.
    We use a Conditional Variational Autoencoder (CVAE)~\cite{kingma2022autoencodingvariationalbayes} to model the human demonstration dataset to reduce the variability of trajectory qualities (See Fig.~\ref{fig:architectures} (c)). 
        
        \subsubsection{Inter-Arm Coordinated Transformer Encoder}
        In this study, we define the encoders so that each robot arm is given as an independent input, including the state of each arm and visual information. The Inter-arm Coordinated transformer Encoder (IACE) captures the synchronization and timing of movements between the arms in a bimanual manipulation task independently of the encoders for each arm, using the state of both arms and visual information that captures the whole scene. The main features are the simultaneous states of both arms and the integration of global information to model the interdependence of the movements of the left and right arms. It is an essential task mechanism that requires precise timing and synchronization between the two arms and allows stable inference of bimanual robot trajectories.
    
        \subsubsection{Multi-arm Decoder}
        One objective of this study is to investigate the structure of the architecture. For the decoder structure, two models are considered: (a) a single decoder and (b) a split decoder as like in Fig.~\ref{fig:architectures}. The single decoder takes a concatenated input of the encoders of the left and right arms, the output tokens of the IACE, and a batch output of the action chunks of the twin arms. Split decoders are structured to process the output of each arm separately, with the decoder associated with each arm concatenating the output tokens of the corresponding encoder and IACE and outputting the action chunks. This research investigates the effectiveness of batch and split outputs for bimanual tasks against these two approaches.

    \begin{figure}[t]
        \centering
        \includegraphics[width=0.98\linewidth]{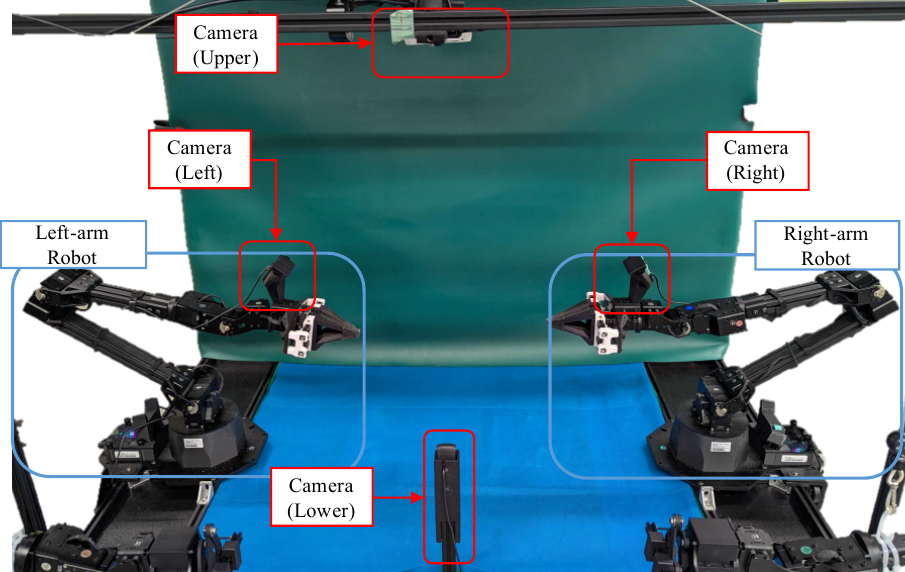}
        \caption{\textbf{Our bimanual robot experimental setting}. It has been designed for real-world applications. We have modified the ALOHA setup for these experiments, including adjusting the camera height and adding a protective tarp around the setup.}
        \label{fig:settings}
    \end{figure}
    
    \subsection{Training and Evaluation}
    The architecture is based on ACT and is trained using an end-to-end imitation learning framework. A human operator collects the dataset employed for training via a teleoperation system, ALOHA~\cite{zhao2023learning} as shown in Fig.~\ref{fig:settings}, encompassing joint positions and RGB images at a rate of 50 Hz. The operators control it slowly to improve precision and data quality. However, during the training phase, the output actions exhibited occasional instability, attributable to minute variations in the joint values. Consequently, the frames were reduced and trained as data with 25 Hz. The camera images were pre-processed to extract joint states and visual features using the ResNet18 backbone~\cite{resnet} to convert the RGB images into feature tokens. Multiple CLS tokens are then added to summarise the information for each segment. Position embedding is incorporated to retain sequence information. The behavior is inferred by employing action chunking~\cite{zhao2023learning}, a technique similar to the base model. A sequence of actions is predicted rather than just a single step in the output value. 

    \begin{figure*}[t]
        \centering
        \includegraphics[width=\linewidth]{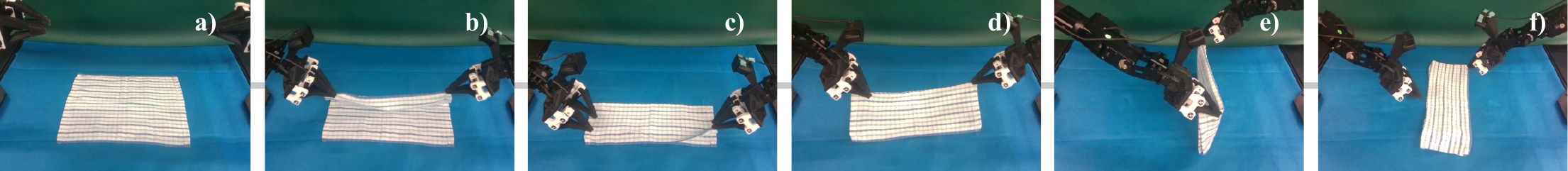}
        \vspace{-15pt}
        \caption{The snapshot of the motion generation in \textbf{Fold Towel} (Asynchronous task). a), b) \& c) is Fold motion, d) is Pick motion, f) is Place motion.}
        \label{fig:exp}
    \end{figure*}

    \begin{figure*}[t]
        \centering
        \includegraphics[width=\linewidth]{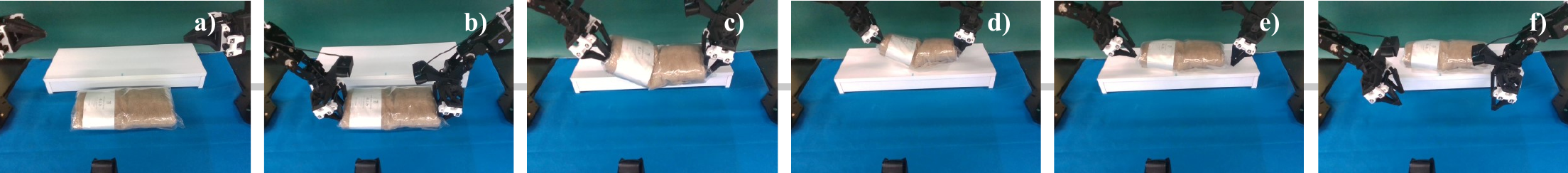}
        \vspace{-15pt}
        \caption{The snapshot of the motion generation in \textbf{Lift and Place Towel}  (Synchronous task). a), b) \& c) are Lift motion, d), e) \& f) are Place motion. }
        \label{fig:exp2}
    \end{figure*}
    
    We employ a temporal ensemble approach during inference, weighting the action predictions over multiple time steps to improve their temporal consistency and robustness~\cite{zhao2023learning}. 

    \begin{table*}[t]
        \centering
        \caption{\textbf{Experimental Results.} Success rate (\%) for four asynchronous tasks (Top) and four synchronous tasks (Bottom), comparing our method with four baselines. The real-world tasks were evaluated over 25 episodes.}
        \includegraphics[width=0.98\linewidth]{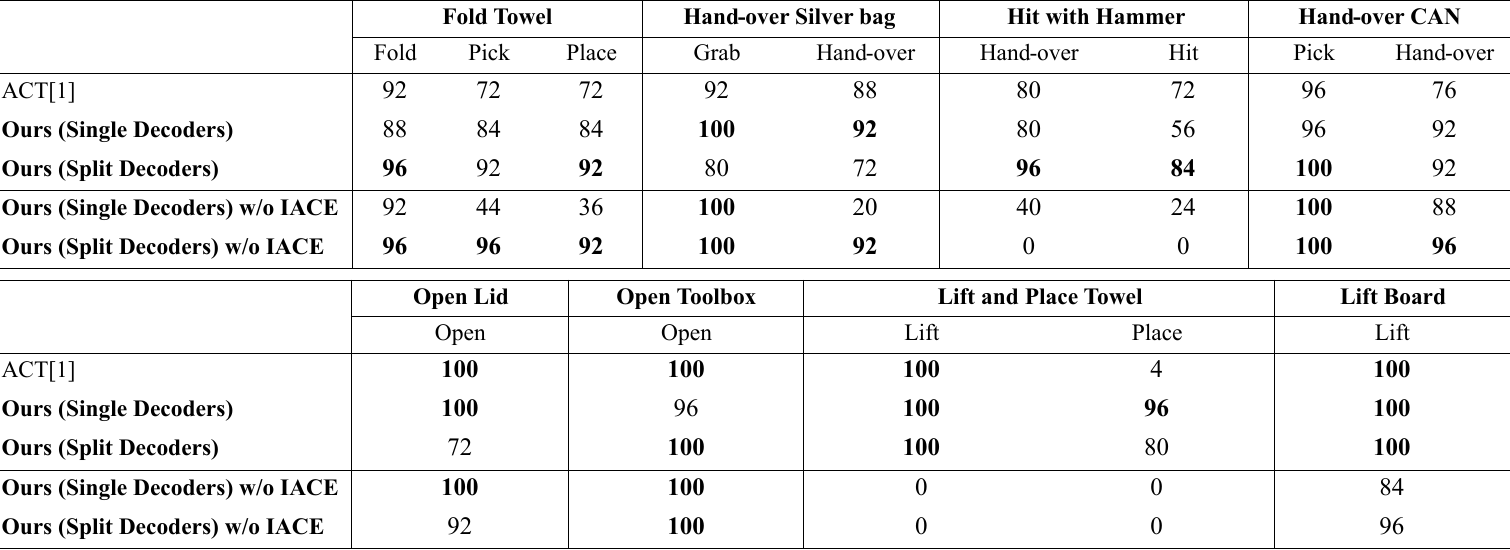}
        \label{tab:results}
    \end{table*}

\section{Experiments and Results}

    We perform our experiments on real-world multi-task datasets, which include teleoperated trajectories by humans. We use the tabletop environment of the ALOHA~\cite{zhao2023learning} setup for the real-robot setup. ALOHA contains two ViperX300S robot arms and four RGB cameras set on each arm wrist (right, left), top view (upper), and horizontal view (lower), as shown in Fig.~\ref{fig:settings}. 

    \subsection{Datasets}

    We design two different types of tasks to evaluate our model, referring to the bimanual taxonomy~\cite{krebs2022bimanual}; asynchronous and synchronous tasks. An asynchronous task is independent of left and right movements in bimanual operations. For example, in grasping an object, the right-hand lifts it first, and then the left-hand grasps it. Conversely, in synchronous tasks, both arms move in sync. We experimented with eight real-world tasks from two types of the bimanual taxonomy: 1) Asynchronous tasks: \textbf{Fold Towel}, \textbf{Hand-over}, \textbf{Silver Bag}, \textbf{Hit with Hammer} and \textbf{Hand-over Can}, 2) Synchronous tasks: \textbf{Open Lid}, \textbf{Open Toolbox}, \textbf{Lift and Place Towel and Lift Board}. 

    In evaluating our proposed models, we compared our model with ACT~\cite{zhao2023learning}. Also, Diffusion Policy is a learning model that uses similarly scaled demonstration data, which is known to be an effective method for learning dexterous tasks. However, as Lee et al.~\cite{lee2024interactinterdependencyawareaction} mentioned, it is not accomplished with just 50 demonstrations. In addition, since our validation aims to investigate the impact of encoder and decoder placement rather than focusing on the detailed learning process, we do not necessarily need to validate other models. Our comparisons are focused solely on ACT, which has already demonstrated superior performance over several different policies in bimanual manipulation tasks.

    \subsection{Implementation Details}
    In our experiments, we acquired 50 episodes on each task. We use the AdamW~\cite{Loshchilov2017DecoupledWD} as the optimizer. We set $\lambda=1\times10^{-5}$. The number of epochs is 5000. We set a batch size of 8. 
    In our training, we used an Ubuntu 20.04 LTS OS PC with a Core i9, 128GB RAM, and a single RTX 4090 GPU (24GB VRAM). The inference time was around 0.02.

    \subsection{Results}
    The snapshots of the motion generation are shown in Fig.~\ref{fig:exp} and Fig.~\ref{fig:exp2}. The results of our experiments are summarized in Tables~\ref{tab:results}. We evaluated the success rates (\%) of our method against four baselines across eight tasks, divided into four asynchronous tasks and four synchronous tasks. The real-world tasks were assessed over 25 episodes. In asynchronous tasks, our method with split decoders, a model architecture where the decoding process is divided into two parts for each arm, consistently outperformed the other baselines in these tasks. In the Fold Towel task, our method with split decoders achieved a success rate of 96\% for picking and placing, compared to ACT's 92\% for picking and 88\% for placing. In the Hand-over Silver Bag task, both ACT and our method with split decoders achieved a perfect success rate of 100\% for grasping and handing over. Our method with single decoders, a model architecture where the decoding process is shared between both arms, surpassed the other baselines in synchronous tasks and demonstrated superior performance. In the Open Lid task, our method (Single decoders) and ACT achieved high success rates. Overall, our method's performance is, on average, higher than that of the other baselines.

    \begin{figure}[t]
        \centering
        \includegraphics[width=0.98\linewidth]{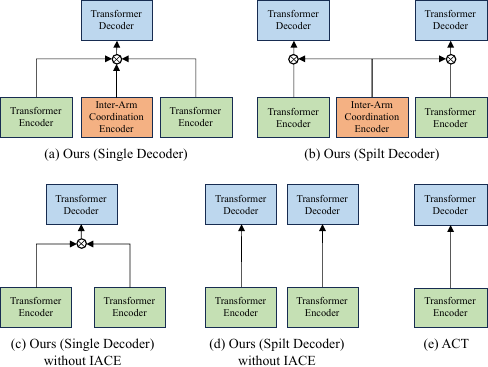}
        \caption{\textbf{Architectures used for verification.} (Top) Our proposed models' architecture includes the architectures with split decoders or single decoder and inter-arm coordinated module. (Bottom) These models do not have the inter-arm coordinated transformer encoder to analyse and compare the model structure with the existing base model.}
        \label{fig:models}
    \end{figure}
    
    \subsection{Ablation Studies}
    We conduct ablation study on various architectures of our models (Fig.~\ref{fig:models}). This study is designed to thoroughly assess the effectiveness of our proposed architecture for general bimanual tasks. We compare our model with split decoders and a single decoder after removing the IACE. The results show that our model's performance declined after the IACE are removed. 


\section{CONCLUSIONS}
This paper presents a robust and adaptable architecture designed to handle various tasks for bimanual robots. The architecture is a key contribution to this research. The encoder and decoder configuration was found to be particularly suitable for bimanual robots, and we introduce the Inter-Arm Coordinated Transformer Encoder (IACE), which synchronizes the movements of both arms, leveraging the policy learning. We perform our proposed models in real-world scenarios by using the teleoperated trajectories from human operators. 

Experimental results showed that the split decoder method consistently achieved higher success rates than other baselines in asynchronous tasks and performed better in complex operations such as towel folding and handover. The single decoder method also achieved better success rates than the baselines.  
It suggests that our module, IACE, was further verified. A significant performance drop was observed when this module was removed, indicating its importance. 
On the other hand, in verifying this study, depending on the type of tasks, the improvement in success rate by the proposed method may not be significant, and improvements in the architecture are required. In the future, we will consider linking the findings from this study with a model compatible with a single arm, further extend it to a multimodal model that integrates text input and other sensor information, and extend it to dual-arm tasks. It is also essential to confirm that there is performance improvement even in methods with different learning algorithms, such as Diffusion Policy~\cite{chi2023diffusionpolicy} and SARNN~\cite{suzuki2023deep}.





\section*{ACKNOWLEDGMENT}

This research was conducted with financial support and using the experimental facilities provided by the National Institute of Advanced Industrial Science and Technology~(AIST). We are grateful for the significant support and assistance provided throughout this study. 


\bibliographystyle{IEEEtran}
\bibliography{references}

\end{document}